# IDRLnet: A Physics-Informed Neural Network Library

Wei Peng[*]    Jun Zhang[*]    Weien Zhou[*]    Xiaoyu Zhao[*]    Wen Yao[*†]
Xiaoqian Chen [*]

**Abstract**

Physics Informed Neural Network (PINN) is a scientific computing framework used to solve both forward and inverse problems modeled by Partial Differential Equations (PDEs). This paper introduces IDRLnet[1], a Python toolbox for modeling and solving problems through PINN systematically. IDRLnet constructs the framework for a wide range of PINN algorithms and applications. It provides a structured way to incorporate geometric objects, data sources, artificial neural networks, loss metrics, and optimizers within Python. Furthermore, it provides functionality to solve noisy inverse problems, variational minimization, and integral differential equations. New PINN variants can be integrated into the framework easily. Source code, tutorials, and documentation are available at `https://github.com/idrl-lab/idrlnet`.

## 1 Introduction

Artificial neural networks are widely applied to solve various scientific computing problems [3, 11, 12, 23]. The Physics-Informed Neural Network (PINN), a deep learning method, is trained simultaneously on both data and the governing differential equations. It is well suited for solving forward and inverse differential equations governing physical systems in various domains such as fluid mechanics [4, 21, 25], solid mechanics [9], and electromagnetic mechanics [16].

Though neural networks have been used as surrogates for solving PDEs since the 1990s [6], this recent increased interest in PINN [20] on engineering and science is due to the increased accessibility of real-world sensor data and open-source platforms such as PyTorch [18] and TensorFlow [1], which provide easy accesses such as high-performance computing and automatic differentiation.

Modeling a large and complex PDE-based physics problem via PINN may make it challenging for new researchers in various fields to use deep learning tools. Several libraries aim to model and solve PDEs through PINN. For example, DeepXDE [15], NVIDIA SimNet [10], SciANN [8], PyDEns [13], Elvet [2], and TensorDiffEq [17] are built on the top of TensorFlow, while NeuroDiffEq [5] is built on the PyTorch; the Julia package NeuralPDE.jl [19] also employs PINN.

---

[*]Defense Innovation Institute, Chinese Academy of Military Science, 100071, Beijing.
[†]Email: wendy0782@126.com

[1]The library is named after the Intelligent Design and Robust Learning (IDRL) laboratory.

This paper introduces an open-source Python package, IDRLnet, developed on PyTorch, designed with physics-informed deep learning. The abstractions used in this programming interface target engineering and scientific applications such as solving differential equations, surrogate training, variational minimization, and parameter identification.

The outline of the paper is as follows. We first describe the primary physics-informed neural networks. We then discuss the central architecture on which IDRLnet relies. Finally, in the example section, we illustrate four applications of IDRLnet, including robust identification of wave equations, incorporating customized adaptive sampling methods, minimizing a variational problem, and solving an integro-differential equation. The examples discussed here and several additional applications are available at https://github.com/idrl-lab/idrlnet.

## 2 Physics-informed neural networks

The primary idea of solving PDEs with neural networks is to reformulate the problem as an optimization problem, where the residual of the differential equations is to be minimized. Without loss of generality, a PDE with initial and boundary conditions can be expressed as

$$\begin{aligned} \mathcal{L}[u](x) &= q(x), & x &\in \Omega \times [0, T], \\ \mathcal{B}[u](x) &= u_b(x), & x &\in \partial\Omega \times [0, T], \\ u(x, 0) &= u_0(x), & x &\in \Omega, \end{aligned} \quad (2.1)$$

where $\mathcal{L}, \mathcal{B}$ are the differential operators, $\Omega$ is the spatio-temporal domain, and $u$ is the solution.

Since the exact solution $u$ usually lies in an infinite dimensional space, one possible numerical method is to parameterize the solution $u$ as $u_\theta$ to approximate the ground truth. In the context of PINN, the model $u_\theta$ is an artificial neural network as the surrogate, where the parameter $\theta$ contains weights and biases of the neural network.

The surrogate $u_\theta(\cdot)$ is a map from $\mathbb{R}^d$ into $\mathbb{R}^N$. By default, IDRLnet employs a simple Multi-Layer Perceptron (MLP) with the activation function $\sigma(\cdot) := swish(\cdot)$ [22], i.e.,

$$u_\theta := L_m \circ \sigma \circ L_{m-1} \circ \sigma \circ \cdots \circ \sigma \circ L_1,$$

where

$$\begin{aligned} L_1(x) &:= W_1 x + b_1, & W_1 &\in \mathbb{R}^{d_1 \times d}, b_1 \in \mathbb{R}^{d_1}, \\ L_i(x) &:= W_i x + b_i, & W_i &\in \mathbb{R}^{d_i \times d_{i-1}}, b_i \in \mathbb{R}^{d_i}, \forall i = 2, 3, \cdots m-1, \\ L_m(x) &:= W_m x + b_m, & W_m &\in \mathbb{R}^{m \times d_{D-1}}, b_D \in \mathbb{R}^m. \end{aligned}$$

If $u_\theta$ is a solution to the differential equation, then the residual terms

$$\begin{aligned} R_{pde} &= \mathcal{L}[u_\theta] - q, & x &\in \Omega \times [0, T], \\ R_b &= \mathcal{B}[u_\theta] - u_b, & x &\in \partial\Omega \times [0, T], \\ R_0 &= u_\theta(\cdot, 0) - u_0, & x &\in \Omega \end{aligned} \quad (2.2)$$

are identically zero.

Unfortunately, since $u_\theta$ is a finite dimension approximation, it is almost impossible for (2.2) equaling 0. One popular way is to incorporate the PDE-induced residuals into the training process of a neural network as components of the loss function. Therefore, the so-called *physics-informed neural networks* are essentially *PDE-informed*, which is the critical idea that IDRLnet bases on, although other kinds of loss functions derived from observed external data or user-defined regularization can also be incorporated.

With a defined loss function $L(\cdot)$, solving the PDE is transformed into the following optimization problem:

$$\min_{\theta} \text{Loss} = \frac{1}{|S_{pde}|} \sum_{x \in S_{pde}} L(R_{pde}(x)) + \frac{1}{|S_b|} \sum_{x \in S_b} L(R_b(x)) + \frac{1}{|S_0|} \sum_{x \in S_0} L(R_0(x)), \quad (2.3)$$

where $S_{pde}$, $S_b$, and $S_0$ are sets of sampling points from the interior domain, the boundary domain, and the initialization domain, respectively. An illustration of a PINN structure is shown in Figure. 1.

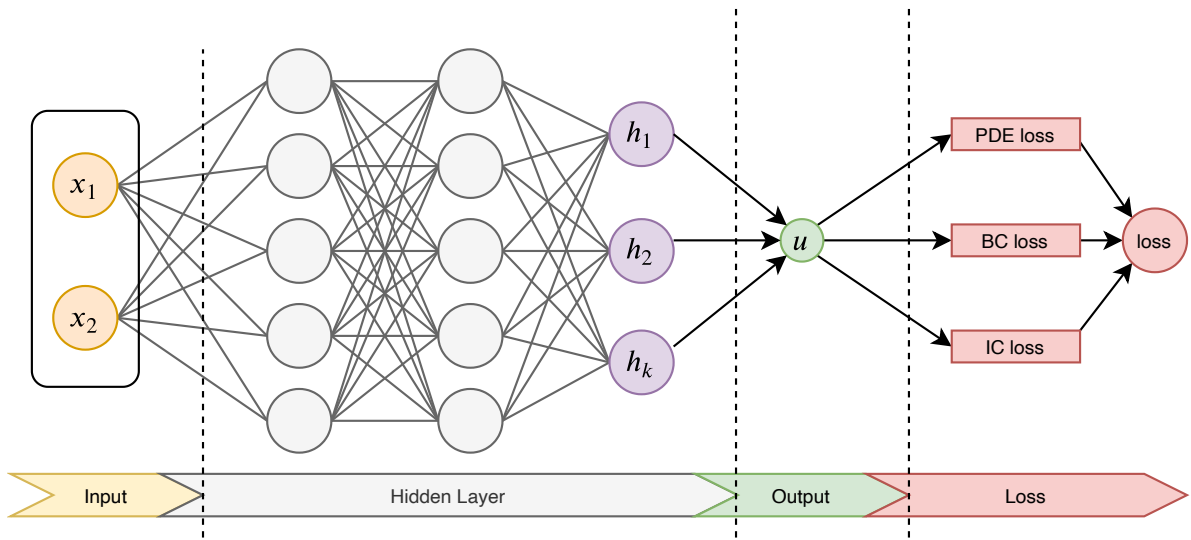

Figure 1: A schematic of a PINN for solving general PDEs.

## 3 Architecture

IDRLnet separates definitions of data sources, PDEs, and neural networks. They are wrapped as nodes with inputs, derivatives, and outputs. In particular, the data sources are wrapped as data nodes with sampling methods implemented; the PDEs and neural networks are wrapped as computational nodes with evaluation methods implemented. A computational node can be evaluated to produce outputs if its inputs and derivatives are accessible from outputs of other nodes.

These nodes are passed as arguments to a solver. The solver links all these nodes according to their types, inputs, derivatives, and outputs. A computational graph is then constructed, and a computational pipeline will be automatically generated according to topological sorting. In the context of supervised learning, each data node samples pairs of variables and responses for training. Each data node corresponds to a sampling domain and a computational pipeline. At the end of these pipelines, the prediction is produced to be compared with the targets. The pairs of prediction values and target values are passed to various loss metrics for producing the final loss. Backpropagation is then performed on the final loss, and an optimizer will try to minimize it.

## 3.1 Geometric Objects

IDRLnet provides several basic geometric objects; signed distance fields of them are also presented. Users can specify sampling density, enveloping bounds, and condition filters. The basic set operations including union(+), subtraction (-), intersection(&) between geometric objects are also available.

The following code illustrates usage of geometric objects.

```
import idrlnet.shortcut as sc
# Define 4 polygons
I = sc.Polygon([(0,0),(3,0),(3,1),(2,1),(2,4),(3,4),(3,5),(0,5),(0,4),(1,4),(1,1),(0,1)])
D = sc.Polygon([(4,0),(7,0),(8,1),(8,4),(7,5),(4,5)]) - sc.Polygon(([5,1],[7,1],[7,4],[5,4]))
R = sc.Polygon([(9,0),(10,0),(10,2),(11,2),(12,0),(13,0),(12,2),(13,3),(13,4),(12,5),(9,5)]) \
    - sc.Rectangle(point_1=(10.,3.), point_2=(12,4))
L = sc.Polygon([(14,0),(17,0),(17,1),(15,1),(15,5),(14,5)])
heart = sc.Heart((18,4), radius=1)
geo = (I + D + R + L + heart)
```

Figure 2 displays the information that the geometric object could provide.

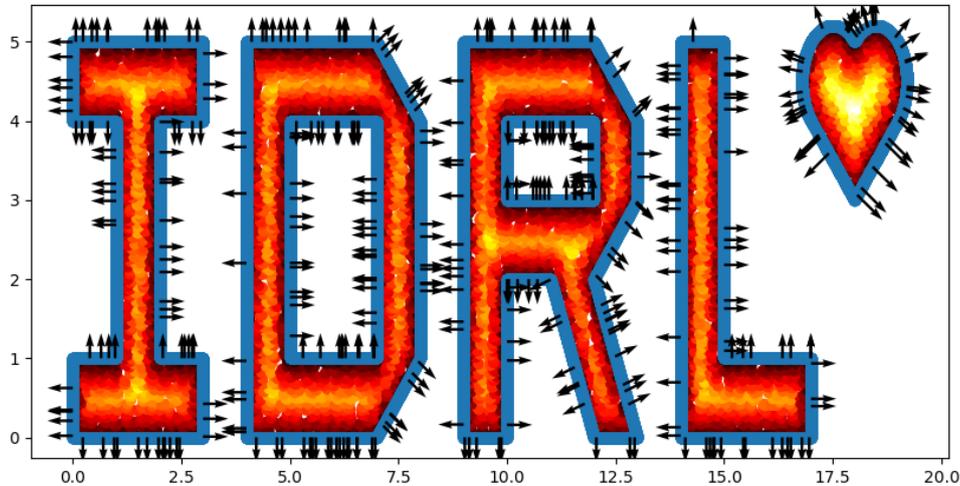

Figure 2: The points on the boundary are colored blue. Arrows on the boundary indicate outward normal directions. The brightness of the interior points indicates the signed distance values.

## 3.2 DataNode

`DataNode` is abstracted from data sources, which provide point collections and the corresponding residual targets(constraints). To illustrate the usage of `DataNode`, we consider an example where points are sampled on both the left and right sides of a square, and the target T is 0 for all these points.

```
import sympy as sp
x, y = sp.symbols('x y')
rec = sc.Rectangle((-1., -1.), (1., 1.))

@sc.datanode
class LeftRight(sc.SampleDomain):
    def sampling(self, *args, **kwargs):
        points = rec.sample_boundary(1000, sieve=((y > -1.) & (y < 1.)))
        constraints = {'T': 0.}
        return points, constraints
```

A schematic diagram is shown in Figure.3. The final loss is composed of components $\{\text{Loss}_i\}$,

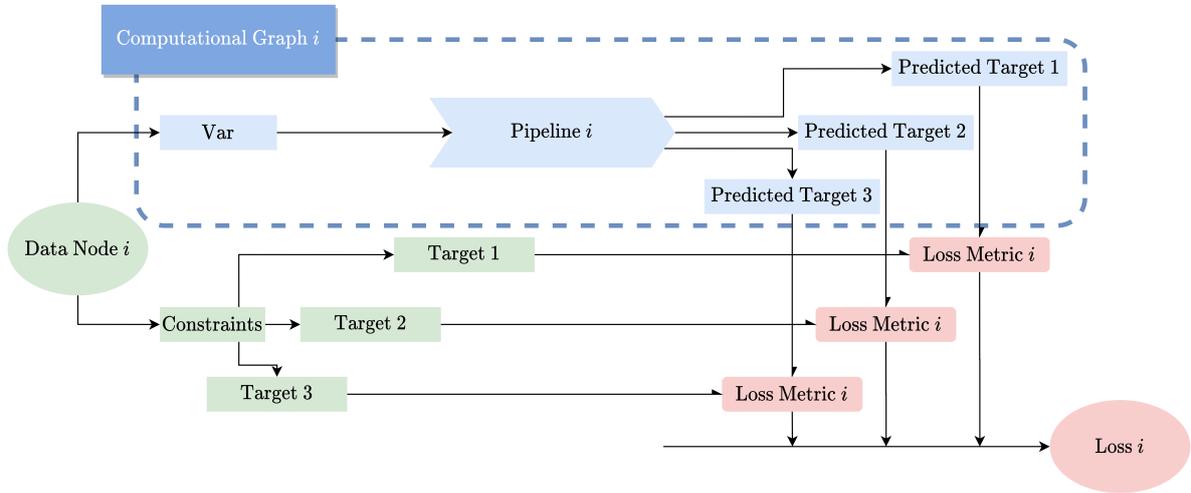

Figure 3: A schematic diagram of one data domain.

$$\text{loss} = \sum_i^M \sigma_i \sum_j^{N_i} \lambda_{ij} \times \text{area}_{ij} \times \text{Loss}_i(\mathbf{y}_j, \mathbf{y}_j^{pred}), \qquad (3.1)$$

where $M$ data domains are included, and the $i$-th domain has $N_i$ sample points in it.

1. By default, the loss function is set to the square loss, and the alternative is the $l_1$ loss. More types will be implemented in the future.

2. $\text{area}_{ij}$ is the weight of the area that the point holds, automatically generated by geometric objects.

3. $\sigma_i$ is the user-defined weight for the $i$-th domain loss, which is set to 1 by default.

4. $\lambda_{ij}$ is the user-defined weight for each point.

## 3.3 Computational Node

IDRLnet has two classes of computational nodes. One is the `NetNode`, the other is the `PDENode`. They are abstracted from neural networks and PDEs, respectively. They both require inputs and necessary derivatives; the outputs will then be obtained after evaluation, which become available as inputs of other computational nodes. The most significant difference between the two classes is whether their instances are trainable. Instances of `NetNode` have trainable parameters while those of `PDENode` do not. IDRLnet has predefined several neural network architectures and PDEs. They are called in the following way:

```
# Define a small-scale MLP
net = sc.get_net_node(inputs=('x', 't',), outputs=('u',),
                                name='net1', arch=sc.Arch.mlp)
# Define a PDE node
pde = sc.BurgersNode(u='u', v=0.01/pi)
# which is equivalent to
pde = sc.ExpressionNode(name='burgers_u',
        expression=u.diff(t)+u*u.diff(x)-0.01/pi*u.diff(x).diff(x))
```

Given one data node and a group of computational nodes, IDRLnet will attempt to generate a pipeline. The initially available inputs of the pipeline are the outputs of the data node. According to the directed connections, computational nodes composes a pipeline to reach the targets (constraints) required by the data node. A schematic is shown in Figure. 4.

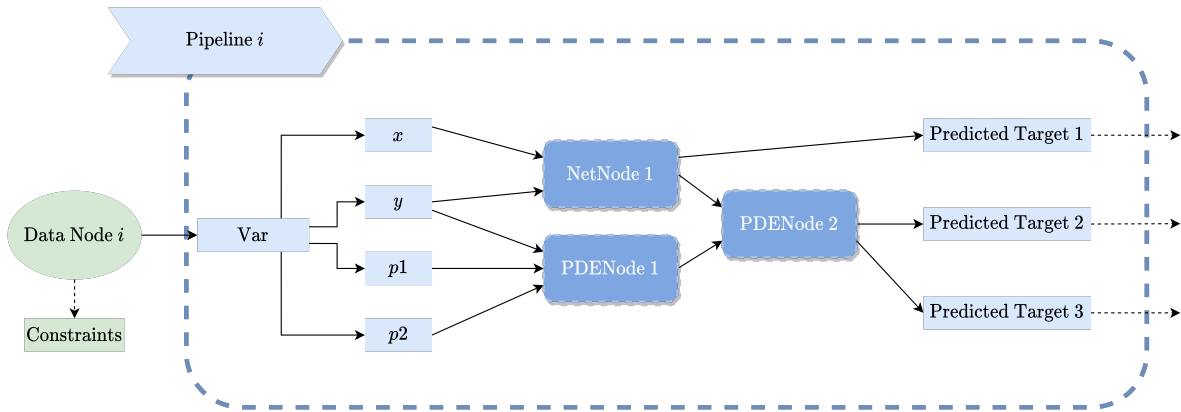

Figure 4: A schematic of one computational graph.

## 4 Examples

To illustrate the applications of IDRLnet, we list four examples in this section. Detailed tutorials, including these four examples, are also available online.

### 4.1 Solving Inverse Noisy Wave Equation

Consider the following wave equation with one space variable,

$$\frac{\partial^2 u}{\partial t^2} = c\frac{\partial^2 u}{\partial x^2},$$

where $c > 0$ is unknown and is to be estimated. A group of data pairs $\{x_i, t_i, u_i\}_{i=1,2,\cdots,N}$ is observed. Then the problem is formulated as:

$$\min_{u,c} \sum_{i=1,2,\cdots,N} \|u(x_i, t_i) - u_i\|^2$$
$$s.t. \frac{\partial^2 u}{\partial t^2} = c\frac{\partial^2 u}{\partial x^2}$$

In the context of PINN, $u$ is parameterized as a neural network $u_\theta$. The problem above is transformed to the discrete form:

$$\min_{\theta,c} w_1 \sum_{i=1,2,\cdots,N} |u_\theta(x_i, t_i) - u_i|^2 + w_2 \sum_{i=1,2,\cdots,M} \left| \frac{\partial^2 u_\theta(x_i, t_i)}{\partial t^2} - c\frac{\partial^2 u_\theta(x_i, t_i)}{\partial x^2} \right|^2. \tag{4.1}$$

The ground truth is
$$u = \sin x \cdot (\sin 1.54t + \cos 1.54t),$$

where $c = 1.54$. We sample about 62 pairs of data from the domain and add outliers by setting $u_i = 3$ to randomized 10 pairs of these samples. IDRLnet defines a network node with a single parameter to represent the undetermined variable $c$.

```
var_c = sc.get_net_node(inputs=('x',),
                        outputs=('c',),
                        arch=sc.Arch.single_var)
```

For robust regression, the $l_1$ loss is usually preferred over the square loss. The conclusion might also hold for inverse PINN as shown in Figure.5.

## 4.2 Adaptive Sampling for Solving Allen-Cahn Equations

This subsection repeats the adaptive PINN method presented by [24] to illustrate that IDRLnet incorporates a user-defined method.

The Allen-Cahn equation has the following general form:

$$\partial_t u = \gamma_1 \Delta u + \gamma_2 \left(u - u^3\right).$$

Consider the one-dimensional Allen-Cahn equation with periodic boundary conditions:

$$u_t - 0.0001 u_{xx} + 5u^3 - 5u = 0, \quad x \in [-1, 1], \quad t \in [0, 1],$$
$$u(0, x) = x^2 \cos(\pi x)$$
$$u(t, -1) = u(t, 1)$$
$$u_x(t, -1) = u_x(t, 1).$$

IDRLnet supports defining user-defined behaviors with the observer pattern [7]. Callback points are shown in Figure. 6.

With the callback functions, every several iterations, points with significant errors are adaptively sampled to increase the weight of the region. Implemented by callbacks provided by IDRLnet, the adaptive sampling procedure is shown in Figure.7.

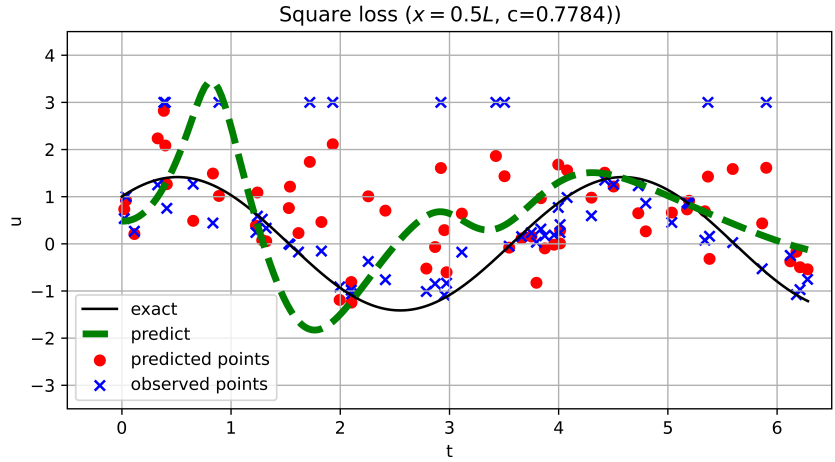

(a) Recovered with the square loss.

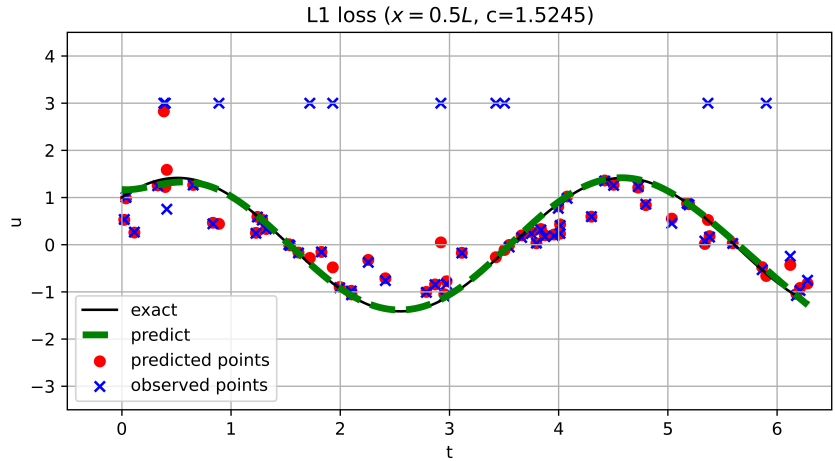

(b) Recovered with the L1 loss.

Figure 5: Note that the points are projected to the u-t plane $x = 0.5L$ so that they may not on the predicted curve. (a) The square loss fails to recover the $c$ such that it fails to reconstruct the solution. (b) Regardless of outliers, the $l_1$ loss successfully reconstructs the solution and estimates $c$ with the absolute error $\Delta c = 0.0155$.

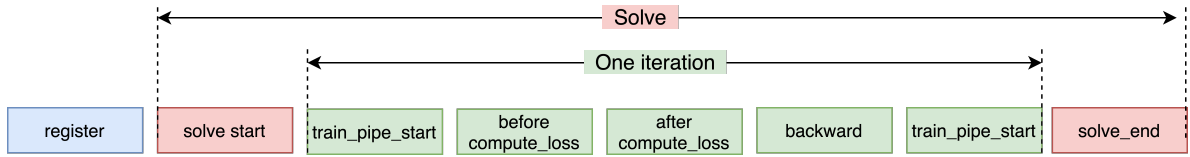

Figure 6: Users can add customized behaviors to implement their methods at these points.

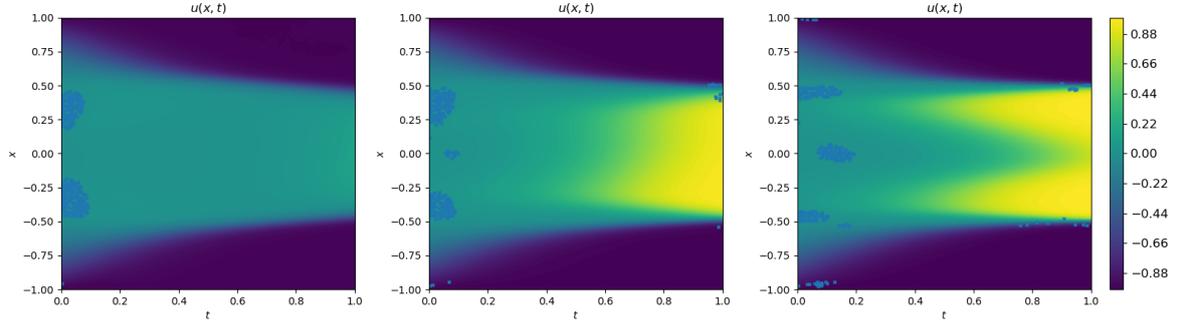

Figure 7: These three figures are drawn at iteration 20000, 30000, and 50000, respectively. The blue points on the figures denote resampled locations.

### 4.3 Solving Volterra integro Differential Equation

We consider an example of the integro-differential equation from [15], the first-order Volterra type integro-differential equation on $[0, 5]$

$$\frac{dy}{dx} + y(x) = \int_0^x e^{t-x} y(t) dt, \quad y(0) = 1, \tag{4.2}$$

with the exact solution $u = \exp(-x) \cosh x$.

The LHS of (4.2) is represented by

```
exp_lhs = sc.ExpressionNode(expression=f.diff(x) + f, name='lhs')
```

The RHS of (4.2) has an integral term with variable limits. Therefore, we introduce the following class to represent the integration:

```
fs = sp.Symbol('fs')
exp_rhs = sc.Int1DNode(expression=sp.exp(s-x)*fs,
                       var=s, lb=0, ub=x,
                       expression_name='rhs',
                        funs={'fs': {'eval': netnode,
                                     'input_map': {'x': 's'},
                                     'output_map': {'f': 'fs'}}},
                       degree=10)
```

We map `f` and `x` to `fs` and `s` in the integral, respectively. The numerical integration is approximated by Gauss–Legendre quadrature with the degree 10. The equivalence between the RHS and the LHS is presented by the difference node,

```
diff = sc.Difference(T='lhs', S='rhs', dim=1, time=False)
```

the output of which is automatically prefixed with `difference_`. The final result is shown in Figure. 8.

### 4.4 Finding a Minimal Surface

Variational problems in engineering are also be solved via constructing neural network surrogates [14]. Solving variational problems are closely related to differential equations. Therefore, IDRLnet can be used to solve variational minimization problems.

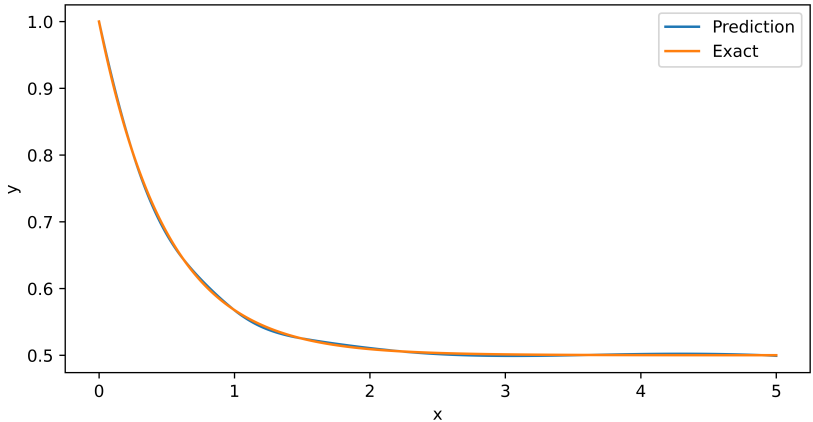

Figure 8: The result of the integro-differential equation. This predicted result matches the exact solution.

To illustrate the usage, we attempt to find a minimal surface of revolution. Given two points $P_1 = (-1, \cosh(-1))$ and $P_2 = (0.5, \cosh(0.5))$, consider a curve $u(x)$ connecting $P_1$ and $P_2$. The surface of the revolution is generated by rotating the curve with respect to the x-axis. This section aims to find the curve that minimizes the surface area. The surface area of the revolution is calculated by integrating over cylinders of radius $u(x)$:

$$S = \int_{x_1}^{x_2} u(x)\sqrt{u'(x)^2 + 1}\,dx. \tag{4.3}$$

IDRLnet supports Monte Carlo integration, which approximates (4.3) and optimizes the estimated value.

IDRLnet also supports loading pre-trained networks. To accelerate convergence, we train the initial network to fit a segment that connects $P_1$ and $P_2$ via IDRLnet. Loading the pretrained model, the solver obtains the result as shown in Figure.9.

## 5 Conclusion

This paper has introduced the open-source deep-learning package IDRLnet, designed to solve PDEs with physics-informed neural networks. It is mainly used for physics-informed deep learning with structured modeling and solving. In a series of examples, we have shown how to use IDRLnet to solve forward and inverse problems for PDEs, solve variational minimization problems and solve integral differential equations. The examples presented here and more examples are available in the Github repository `github.com/weipeng0098/idrlnet`.

IDRLnet is based on the PyTorch package. However, the high-performance computing capabilities, including CPU/GPU parallelization, have not been fully exploited. Addressing the issue will be one of the future development directions of IDRLnet.

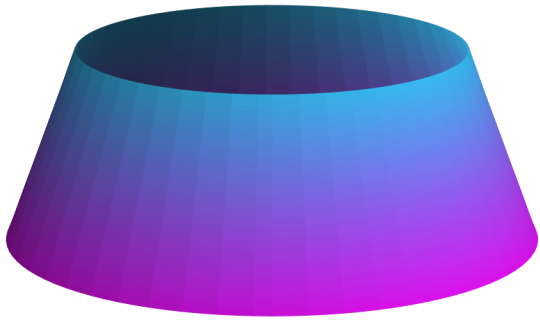

(a) Iteration=0

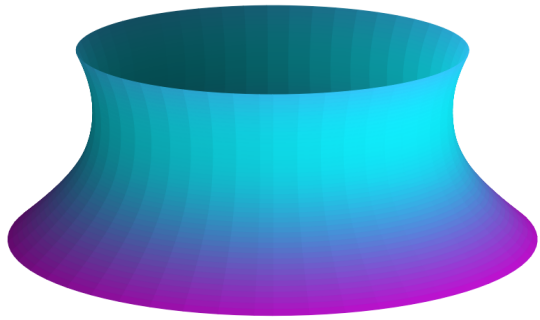

(b) Iteration=400

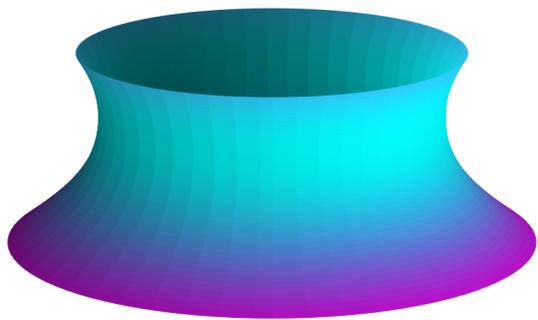

(c) Iteration=800

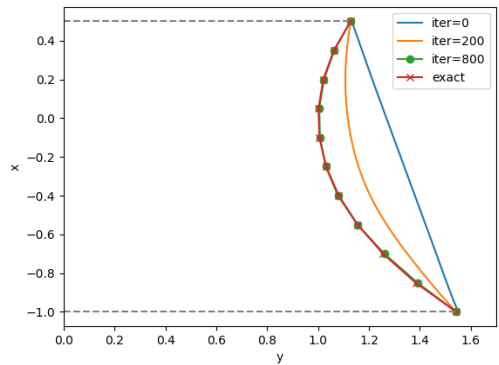

(d) Iterations

Figure 9: (a) Due to our pre-trained initialization, the training process starts with a truncated cone. (c)(d) After 800 iterations, the result converges to a catenoid.

# Acknowledgements